# Deep Learning Training with Simulated Approximate Multipliers

Issam Hammad, Kamal El-Sankary, and Jason Gu

Electrical and Computer Engineering Department, Dalhousie University, Halifax, NS, Canada, B3H 4R2

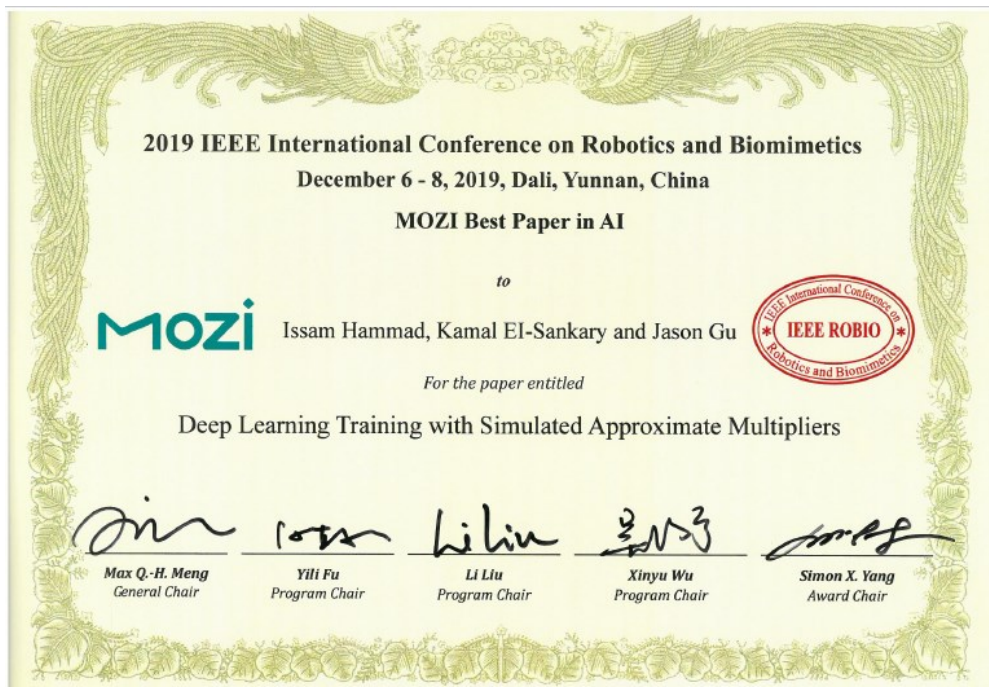

**For Citation Please Use:**

Issam Hammad, Kamal El-Sankary, and Jason Gu. "Deep Learning Training with Simulated Approximate Multipliers." *2019 IEEE International Conference on Robotics and Biomimetics (ROBIO)*. IEEE, 2019. (Pages: 47 - 51)

**IEEE URL:** https://ieeexplore.ieee.org/document/8961780

# Deep Learning Training with Simulated Approximate Multipliers

Issam Hammad, IEEE Student Member, Kamal El-Sankary, IEEE Member, and
Jason Gu, IEEE Senior Member

Electrical and Computer Engineering Department, Dalhousie University, Halifax, NS, Canada, B3H 4R2
Email: issam.hammad@dal.ca

*Abstract*— **This paper presents by simulation how approximate multipliers can be utilized to enhance the training performance of convolutional neural networks (CNNs). Approximate multipliers have significantly better performance in terms of speed, power, and area compared to exact multipliers. However, approximate multipliers have an inaccuracy which is defined in terms of the Mean Relative Error (MRE). To assess the applicability of approximate multipliers in enhancing CNN training performance, a simulation for the impact of approximate multipliers error on CNN training is presented. The paper demonstrates that using approximate multipliers for CNN training can significantly enhance the performance in terms of speed, power, and area at the cost of a small negative impact on the achieved accuracy. Additionally, the paper proposes a hybrid training method which mitigates this negative impact on the accuracy. Using the proposed hybrid method, the training can start using approximate multipliers then switches to exact multipliers for the last few epochs. Using this method, the performance benefits of approximate multipliers in terms of speed, power, and area can be attained for a large portion of the training stage. On the other hand, the negative impact on the accuracy is diminished by using the exact multipliers for the last epochs of training.**

## I. INTRODUCTION

With the accelerated increase in computational power, cloud computing resources, and the availability of data, deep learning [1] has become a viable approach to solve artificial intelligence problems in various fields. Deep learning is used nowadays in many fields and applications such as self-driving cars, image recognition and classifications, robotics, health care, and security. One of the major challenges that deep learning faces is the slow training time especially using very deep neural networks with enormous data to train. Training a deep convolutional neural network usually requires thousands of feed-forward and back-propagation iterations. In each iteration, all the network weights are updated. These weights can be in millions as in the case of the VGGNet-16 network in [2] which has 138M weights. The primary mathematical operation in a deep convolutional network is multiplication, therefore, any reduction in the cost of the multiplication will lead to a major enhancement to the performance of the entire system.

Approximate computing provides a solution to enhance performance in terms of speed, power, and area at the cost of a pre-defined error range in the obtained output. One of the primary applications for approximate computing is the approximate multipliers. Several approximate multiplier designs were proposed in the literature such as [3-6]. Using these multipliers can lead to significant performance enhancements. However, these enhancements have a cost of inaccuracy in the output which is usually defined by the Mean Relative Error (MRE). As an example, the multiplier in [3] achieves performance enhancements of 47% in speed, 50% in area, and 59% in power. However, these enhancements have a cost of an inaccuracy in the output defined by a Gaussian MRE of 1.47% with Standard Deviation (SD) of 1.803%. Like [3], the approximate multiplier designs [4-6] have different performance enhancements with a predefined MRE error. The MRE is defined in equation (1), where, $Xi$ is the exact multiplication value, $Xi'$ is approximate multiplication value from an approximate multiplier, and n is the total number of samples.

$$MRE = \frac{1}{n}\sum_{i=1}^{n}\frac{|Xi' - Xi|}{|Xi|} \qquad (1)$$

In a previous work [7], we have studied the impact of using approximate multipliers on the inference stage of a pre-trained CNN network. The simulated MRE and SD in [7] were selected to approximately simulate the reported inaccuracies by various approximate multipliers in the literature such as [3-6]. The work in [7] has demonstrated that with minimal cost of added inaccuracy, approximate multipliers can be used to enhance to significantly boost the inference performance of a pre-trained deep convolutional neural network (CNN) in terms of speed, power, and area.

Lower the training cost in terms of the power, area, and speed can be very beneficial in the case of training on the edge. Training is computationally very expensive, and it is usually performed using high powerful servers. However, in certain instances, training will have to be performed at least partially on the edge. When training on the edge is performed, the model will be trained partially or fully using the low power embedded hardware.

Fully autonomous mobile robots with image classification abilities are a perfect example of the need for training on the edge at least partially. These robots are used in various industries today such as aerospace applications, nuclear power plants, oil refineries, chemical factories, underwater and military applications. In many instances, these robots will be operating in offline areas without any connection to the main server. Therefore, for the purpose of improving prediction accuracy, continuous model training could be required and will have to be performed on the edge. Hence, as a result of this need for deep learning train the edge, proposing methods to improve training performance in terms of power, area, and speed becomes vital. This can be achieved by utilizing approximate multipliers during training as this paper will present.

This paper's objective is to propose new methods to enhance the training stage performance for a deep CNN by simulating the impact of the approximate multiplier error during training. One of the primary research contributions is a new training methodology which enhances the training performance of deep convolutional networks without any negative impact on the accuracy. This is accomplished by training the network using two phases. In the first phase, the training starts using approximate multipliers, then in the second phase the training switches to exact multipliers. Using this methodology all the performance gains of approximate multipliers can be obtained during the first phase of the training, while in the second phase, any negative impact on the accuracy caused by the approximate multipliers will be diminished. The number of iterations for each phase is a variable that is determined by the inaccuracy of the approximate multiplier.

This paper is structured as follows, in section II presents the details of the used deep CNN and dataset. Section III demonstrates the achieved inference accuracy by training the deep CNN with simulated approximate multiplier error. Section IV presents the new proposed hybrid training approach. In Section V the research conclusions are summarized.

## II. SIMULATION ENVIRONMENT

For the simulation, a modified version of the VGGNet was used as a model. This modified version was proposed by [8] and it slightly differs from the original design which was proposed by [2]. The design in [8] is tailored to work with CIFAR-10 image dataset [9] which is used in this study. The model in [8] is smaller than the original model in [2] as it has a 32x32 input size rather than a 224x224 input, additionally, it has 2 fully dense layers rather than 3 fully dense layers. It also includes batch normalization and dropout to reduce overfitting. Figure 1 demonstrates the modified architecture of the VGGNet which is used in for the simulations in this paper. Note that in Figure 1, the first two numbers in the brackets contain the image dimension while the third number reflects the number of filters. CIFAR-10 dataset [9] consists of 60000 color images divided equally into 10 classes with 50000 images used for training and 10000 images used for testing.

For development, the popular deep learning Python library Keras was used [10]. The used model was adopted from the repository in [11], which presents an implementation of the design proposed in [8]. In this paper, the repository in [11] was modified to evaluate the applicability of training with approximate multipliers. Table I specifies the used training configurations for all the test cases in this paper. During the simulation, the type "float16" was used. To simulate the impact of approximate multipliers on the training, the multiplication should have an inaccuracy defined by a certain MRE and SD. This was implemented using a Keras custom layer functionality. These layers were added before every convolutional and dense layer. These custom layers were programmed to mimic the impact of the error in approximate multipliers by creating a multiplication inaccuracy based on a specific MRE and SD during both backpropagation and forward propagation. These layers simulate this inaccuracy through elementwise multiplication between the weights and a generated error matrix. Each network layer had a unique error matrix which simulated a certain MRE and SD. Having these custom layers throughout the network simulated the impact of the approximate multiplier error on the overall accuracy.

TABLE I
TRAINING CONFIGURATIONS

| Parameter | Value/Method |
| --- | --- |
| Epochs | 200 |
| Batch Size | 128 |
| Output Classes | 10 |
| Activation Function | ReLU |
| Loss Function | Categorical crossentropy |
| Optimizer | Stochastic gradient descent (SGD) optimizer with learning rate decay |
| Dataset | CIFAR-10 |
| Training Images | 50000 |
| Testing Images | 10000 |
| Regularization | L2 Regularization with weight decay (0.0005) and Dropout of 30%-50% |
| Normalizaion | Input Normalization and Batch Normalization |

Figure 2 illustrates a histogram of a sample error matrix which is used to simulate an MRE of ~3.6% and an SD of ~4.5%. Many of the reported approximate multipliers MREs have a near zero-mean Gaussian distribution, this can be seen in the approximate multipliers [3] and [4]. Therefore, to provide a generic simulation that can be applicable to many approximate multiplier models, all the simulated MRE values in this paper were based on a near zero-mean Gaussian distribution.

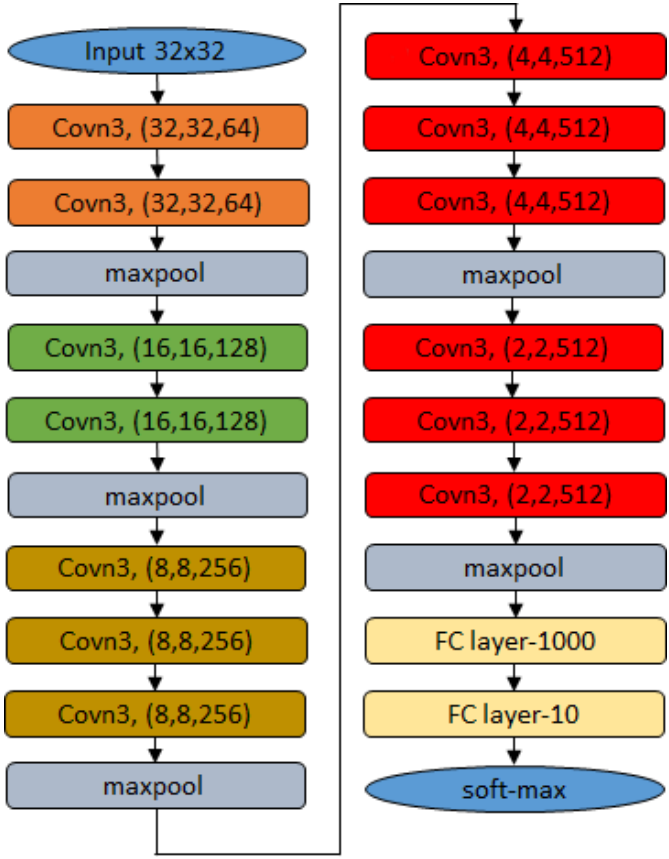

Figure 1. Modified VGGNet architecture which was used for this study

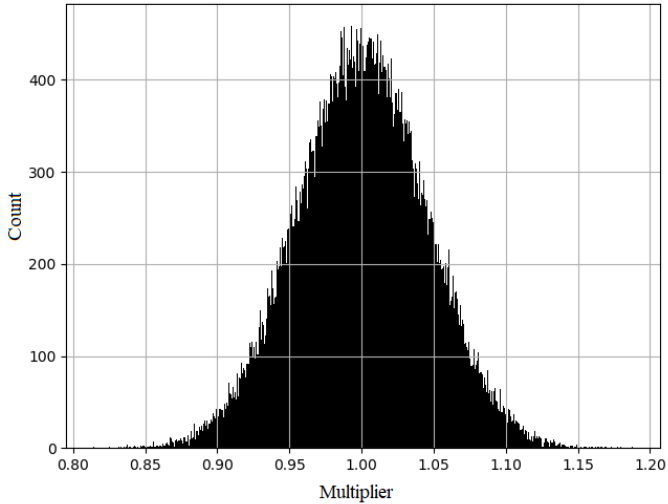

Figure 2. A histogram (500 bins) of a sample error matrix (MRE=~3.6%, SD=~4.5%)

## III. TRAINING WITH SIMULATED APPROXIMATE MULTIPLIER ERROR

As described in the previous section, simulating the approximate multiplier error during the training stage was achieved using Keras custom layers. These layers create a multiplication inaccuracy based on the tested MRE and SD. The error simulation is achieved by multiplying the layers' weights with an error matrix which is generated to simulate the desired MRE.

Figure 3 demonstrates the followed process for this simulation. After loading the data, an error matrix with an approximate MRE and SD was generated for each layer. This simulated the impact of an approximate multiplier on the accuracy of the network. By using Keras custom layers as described in the previous section, the approximate multiplier error simulation was applied during all forward propagation and backpropagation iterations. During the training, the weights after certain training epochs were downloaded. This allowed the training to resume from that epoch when reloading the model, also this was needed to implement the hybrid approach which will be discussed in the next section. After completion, the final model weights were downloaded, and the model was reloaded for testing purposes. The testing stage excluded the simulation of the approximate multiplier error, therefore, all the added Keras custom layers were removed. This ensured that any impact on the inference accuracy is resulting from applying the approximate multipliers simulation during the training stage only.

Table II presents the achieved inference accuracy as a result of training with simulated approximate multiplier error. The table lists the results for different approximate multiplier configurations based on their MRE and SD. The first row presents the inference accuracy achieved as a result of training with an exact multiplier which excludes any error simulations. This achieved accuracy will be referred to as the baseline accuracy (93.6%). The remaining cases are reported based on the simulated MRE and SD values. In each test case, the achieved inference accuracy and the difference in accuracy compared to the baseline accuracy are reported. The simulation differences between the presented test cases were limited to the ranges of MRE and SD, this guaranteed a fair performance comparison among the presented test cases. As can be seen from the table the impact of the approximate multiplier error during training on the achieved inference is very small, especially for lower MRE cases.

To clearly demonstrate the benefit of this simulation, a mapping can be done between the simulated test cases and the reported performances of approximate multipliers in the literature. For example, DRUM [3] reported performance enhancements of 47%, 50% and 59% in the speed, area, and power, respectively with a cost of a near zero-mean Gaussian distribution with MRE=1.47% and SD=1.803%. This is very close to test case 2 in Table II with MRE=~1.4% and SD=~1.8% which also has a zero-mean Gaussian distribution. In other words, using the approximate multiplier DRUM [4] in a custom design can approximately accelerate all the multiplications of the network during training by 47% with a cost of a drop in the inference accuracy by only 0.07%.

Any improvement in the multiplication performance will directly boost the convolution performance as convolution is just a series of Multiplication and Accumulation (MACs) operations. Based on [12], the convolution in a CNN consumes 90.7% of the total computational time required by the network.

Thus, any performance improvement on the multiplication will directly affect the performance of the entire network.

Based on [13], there is a high correlation between the approximate multiplier error and the performance gains achieved. Hence, using approximate multipliers with higher error leads to higher performance gains for a custom hardware design for CNN training. Nevertheless, the network's ability to tolerate error is limited, after a certain level the network accuracy will collapse. This can be seen in the significant drop in accuracy in test cases 7 and 8 in Table 2. Therefore, a balance must be maintained between the approximate multiplier error and the CNN performance gains.

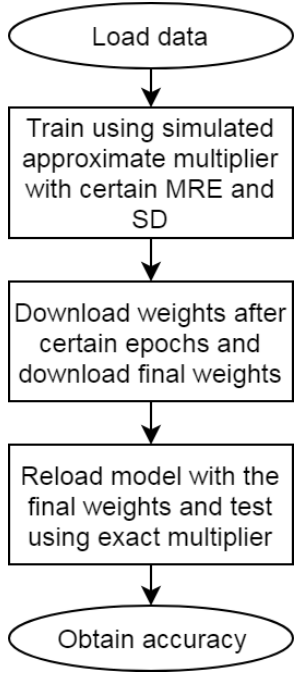

Figure 3. The followed procedure for simulating the impact of approximate multipliers on the training stage

TABLE II
INFERENCE ACCURACY BASED ON TRAINING WITH SIMULATED APPROXIMATE MULTIPLIER ERROR

| Test ID | MRE | SD($\sigma$) | Achieved Accuracy | Diff. From Exact |
|---|---|---|---|---|
| 0 | 0% | 0% | 93.6% | N/A |
| 1 | ~1.2% | ~1.5% | 93.59% | -0.01% |
| 2 | ~1.4% | ~1.8% | 93.53% | -0.07% |
| 3 | ~2.4% | ~3.0% | 93.35% | -0.25% |
| 4 | ~3.6% | ~4.5% | 93.23% | -0.37% |
| 5 | ~4.8% | ~6.0% | 93.11% | -0.49% |
| 6 | ~9.6% | ~12% | 93% | -0.60% |
| 7 | ~19.2% | ~24% | 92.23% | -1.37% |
| 8 | ~38.2% | ~48% | 65.65% | -27.95% |

## IV. THE HYBRID TRAINING APPROACH

As presented in the previous section, despite the great performance gains that can be achieved by training with approximate multipliers, a cost of a slight drop in the network accuracy is inevitable. To eliminate this cost, a hybrid training methodology can be applied which involves using both approximate multipliers and exact multipliers. Using this methodology, the training starts with approximate multipliers then switches to exact multipliers for the last epochs of the training By evaluating this hybrid training methodology, test cases 1-6 in Table II and up to MRE=~9.6% reached an accuracy within 0.02% of baseline accuracy.

Table III illustrates, the number of epochs that were used by the approximate multipliers then the exact multipliers to achieve an inference accuracy which is equal or greater than 93.58% (0.02% less than the baseline accuracy).

In deep learning, neural network weights can be downloaded at any point and the training can be resumed from pre-loaded weights. Therefore, realizing the hybrid approach using a custom hardware design is not complicated. For example, one chip can be designed for training using approximate multipliers and the other using exact multipliers, the exact multiplier training chip can resume and finish what was partially training by the approximate multiplier training chip.

TABLE III
HYBRID TRAINING CONFIGURATIONS FOR DIFFERENT MRE VALUES

| Test ID | MRE | Appr. Multiplier Epochs | Exact Multiplier Epochs | Approximate Multiplier Utilization |
|---|---|---|---|---|
| 1 | ~1.2% | 200 | 0 | 100% |
| 2 | ~1.4% | 191 | 9 | 95.5% |
| 3 | ~2.4% | 180 | 20 | 90% |
| 4 | ~3.6% | 176 | 24 | 88% |
| 5 | ~4.8% | 173 | 27 | 86.5% |
| 6 | ~9.6% | 151 | 49 | 75.5% |

The results in Table III were obtained by following the procedure presented in the flowchart in Figure 4. Table III presents the optimal hybrid solution found for each test case and. In this procedure, the training started by loading partially trained model weights from a simulated approximate multiplier up to certain epoch. These weights were saved after certain epochs during the simulations which were presented in the previous section. After loading these partially trained models, the remainder of the training was resumed by an exact multiplier up to 200 epochs as specified in Table I. Finding this optimal solution required tuning the switching epoch between the approximate and the exact multiplier increasing it or decreasing it until finding the optimal combination.

Table III presents the optimal hybrid solution for this hybrid approach. Nevertheless, in production, any repeat in training must be avoided as it defeats the purpose of performance enhancement. Subsequently, it will be challenging to obtain this optimal switching epoch prior to training and without computational costs. However, using a non-optimal solution by approximating the switching epoch index for the hybrid approach still achieves significant performance gains. If the final achieved inference accuracy by the hybrid approach is almost equal to the exact multiplier accuracy, any utilization of the approximate multipliers for the initial epochs are pure performance gain with almost no cost.

In general, developers usually keep training until there are no further improvements to the cross-validation accuracy. Therefore, regardless of what the initial switching epoch index was, the target accuracy can be achieved if the approximate multiplier error is suitable for the application. In the case of starting with an initial switching epoch index less than the optimal, the target accuracy should be achieved by the final epoch. On the other hand, if the initial switching epoch was larger than the optimal, the target accuracy can be achieved by training for a few additional epochs. In both cases, the norm is to keep training until the cross-validation accuracy flattens. Therefore, the advantage of training the initial epochs with approximate multipliers can be attained regardless.

a certain epoch. The simulation results of the hybrid approach show that using exact multipliers for the last epochs can eliminate any accuracy drop caused by the usage of approximate multipliers initially. Therefore, significant performance gains can be achieved by utilizing approximate multipliers for a large portion of the training while having almost no negative impact on the final achieved accuracy.

While this concept can be used to enhance the performance of deep learning training in general, it is particularly beneficial in the case of training on the edge. Training on the edge is required in the case of offline systems such as in the case of offline mobile robots performing in remote harsh environments. Lowering the edge training cost for these robots in terms of power, area, and speed is vital for performance. This performance enhancement can be achieved using approximate multipliers for the training with minimal impact on the accuracy as the paper presented.

## VI. ACKNOWLEDGEMENT

This work was supported by The Killam Trusts Scholarship.

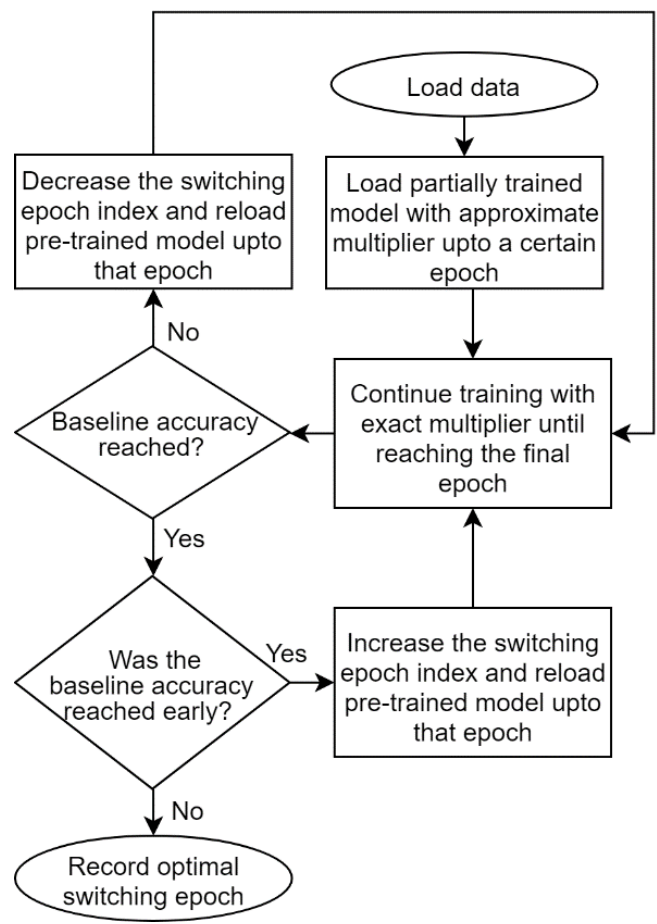

Figure 4. The followed procedure for finding the optimal solution for the hybrid training approach

## V. CONCLUSION

In this paper, the concept of utilizing approximate multipliers to enhance the training performance of deep CNN was proposed. Simulation results show that using approximate multipliers for CNN training result in a minimal drop in accuracy while having the potential to achieve significant performance gains in custom hardware designs. Additionally, a hybrid approach was proposed in which the training starts approximate multipliers then switches to exact multipliers after

REFERENCES

[1] LeCun, Yann, Yoshua Bengio, and Geoffrey Hinton. "Deep learning." nature 521.7553 (2015): 436.

[2] Simonyan, Karen, and Andrew Zisserman. "Very deep convolutional networks for large-scale image recognition." arXiv preprint arXiv:1409.1556 (2014).

[3] Hashemi, Soheil, R. Bahar, and Sherief Reda. "DRUM: A dynamic range unbiased multiplier for approximate applications." Proceedings of the IEEE/ACM International Conference on Computer-Aided Design. IEEE Press, 2015.

[4] Leon, Vasileios, et al. "Approximate Hybrid High Radix Encoding for Energy-Efficient Inexact Multipliers." IEEE Transactions on Very Large Scale Integration (VLSI) Systexact multipliers 26.3 (2018): 421-430.

[5] Venkatachalam, Suganthi, and Seok-Bum Ko. "Design of power and area efficient approximate multipliers." IEEE Transactions on Very Large Scale Integration (VLSI) Systexact multipliers 25.5 (2017): 1782-1786.

[6] Yang, Tongxin, Tomoaki Ukezono, and Toshinori Sato. "Low-Power and High-Speed Approximate Multiplier Design with a Tree Compressor." Computer Design (ICCD), 2017 IEEE International Conference on. IEEE, 2017.

[7] Hammad, Issam, and Kamal El-Sankary. "Impact of Approximate Multipliers on VGG Deep Learning Network." IEEE Access 6 (2018): 60438-60444.

[8] Liu, Shuying, and Weihong Deng. "Very deep convolutional neural network based image classification using small training sample size." Pattern Recognition (ACPR), 2015 3rd IAPR Asian Conference on. IEEE, 2015.

[9] A. Krizhevsky and G. Hinton, ''Learning multiple layers of features from tiny images,'' University of Toronto, Toronto, ON, Canada, Tech. Rep.,2009. [Online]. Available: https://www.cs.toronto.edu/~kriz/learning-features-2009-TR.pdf

[10] Chollet. (2015). Deep Learning for Humans. [Online]. Available:https://github.com/fchollet/keras

[11] Y. Geifman.VGG16 Models for CIFAR-10 and CIFAR-100 Using Keras. Accessed: May 2017. [Online]. Available: https://github.com/geifmany/cifar-vgg

[12] Cong, Jason, and Bingjun Xiao. "Minimizing computation in convolutional neural networks." International conference on artificial neural networks. Springer, Cham, 2014.

[13] Mrazek, Vojtech, et al. "Design of power-efficient approximate multipliers for approximate artificial neural networks." Proceedings of the 35th International Conference on Computer-Aided Design. ACM, 2016.